\documentclass[11pt]{article}

\usepackage[utf8]{inputenc}
\usepackage{microtype}
\usepackage[margin=2.5cm]{geometry}
\usepackage{amsmath}
\usepackage{enumitem}
\setlist{itemsep=2pt,topsep=4pt}
\usepackage[numbers,sort&compress]{natbib}
\usepackage{xurl}
\usepackage[dvipsnames]{xcolor}

\usepackage[colorlinks=true,linkcolor=blue!50!black,citecolor=blue!50!black,urlcolor=blue!50!black]{hyperref}
\hypersetup{
  pdftitle={Why and How to Do Science When AI Is Better at It: A Framework Grounded in Human Understanding and Direction},
  pdfauthor={Hoda Heidari and Aydin Mohseni}
}

\title{Why and How to Do Science When AI Is Better at It:\\ A Framework Grounded in Human Understanding and Direction}
\author{Hoda Heidari\\[2pt]
  \small Machine Learning Department and Software and Societal Systems Department\\
  \small Carnegie Mellon University, Pittsburgh, PA, USA\\
  \small \href{mailto:hheidari@andrew.cmu.edu}{\texttt{hheidari@andrew.cmu.edu}}
  \and Aydin Mohseni\\[2pt]
  \small Department of Philosophy\\
  \small Carnegie Mellon University, Pittsburgh, PA, USA\\
  \small \href{mailto:amohseni@cmu.edu}{\texttt{amohseni@cmu.edu}}}
\date{}

\begin{document}
\maketitle

\begin{abstract}
As AI begins to match or exceed human researchers at core scientific tasks, two questions become pressing: \emph{why} should humans keep doing science, and \emph{how} should they use AI when they do? This paper argues that the second can be answered only by answering the first. It holds that one indispensable aim of science is \emph{durable, collective human understanding of the world, pursued in service of human ends and values}. From this principle it derives a two-part test for when a research task can be delegated to AI. The first part, \emph{human understanding}, requires that the scientific community remain able to explain, defend and contest the resulting claims, and retain the skills to do so. The second, \emph{human direction}, requires that judgments about which questions matter and what findings mean stay with accountable humans. Applying the test to each phase of research in the Contributor Roles Taxonomy (CRediT) yields recommendations that fields can tailor to their questions and methods. The framework differs from existing guidance in three ways. It rests on an account of what science is for rather than on integrity norms. It holds even if AI outperforms human researchers, rather than depending on AI's current limitations. And it judges each use of AI by its effects on human understanding and direction rather than by the type of tool, across every phase of research, including peer review and the training of new researchers. It closes with the reforms needed to keep science aimed at human understanding and direction.
\end{abstract}

\noindent\textbf{Keywords:} artificial intelligence, scientific understanding, philosophy of science

\section{Introduction}\label{sec:intro}

Imagine an oracle of knowledge. Ask it any question and it returns a correct answer: a vaccine against a new virus, a machine-checkable proof of a long-open conjecture, the structure of a protein no one has yet imaged. It does not tell us why its answers are correct, and we may be unable to work this out for ourselves. Most of us would use its answers, and we should. But would there still be a reason for humans to do science? Someone would still have to decide which questions to put to the oracle and what its answers mean for us. And a world that relied on it would be one in which no one could explain why the vaccine works, or carry on if the oracle were withdrawn. This paper argues that such a world would have knowledge but would have lost something science exists to provide, and that the case for human science rests on what that something is.

This world is less hypothetical than it sounds. AI systems now generate hypotheses, run analyses and write papers, and recent results suggest they may come to do so better and faster than human researchers \cite{lu2026,mitchener2025}. This raises two questions for researchers. The first is philosophical: \emph{why} humans should keep doing science at all if AI can do it better. The second is practical: \emph{how} to incorporate AI in research without giving up ownership of the work. We argue that the second question can be answered only by answering the first, and that answering the first requires returning to why we do science.

Our answer is that one indispensable aim of science is \textbf{durable, collective human understanding of the world, pursued in service of human ends and values}. Knowledge that exists somewhere, or that can be put to use, does not by itself meet that aim. It follows that research should remain human-led to preserve this principle even when AI is more capable at producing reliable knowledge and instruments to act in the world, and that AI can be used wherever its use does not erode the principle.

We argue under a deliberately strong assumption: that AI is more accurate, less biased and more efficient than human researchers at the core tasks of science.\footnote{The assumption concerns AI's competence at the tasks of science, not its alignment with human values. We do not assume that AI will act on our values, and the argument does not depend on its failing to. Even an AI that could predict perfectly what people value would not settle what they should pursue, because human ends and values are not only outcomes to be forecast; they are formed and legitimated through a process. We return to this point in Section~\ref{sec:obj4}.} The argument therefore does not depend on the current flaws of AI. If human-led research is justified in that world, it is justified in ours. The central danger in such a world is that humanity may become increasingly capable of producing and using scientific knowledge while becoming progressively less capable of understanding, contesting, and directing it. We posit that in this world, humans should continue to participate in science not because they are necessarily the best performers of scientific tasks, but because scientific inquiry is one of the main institutions through which humanity develops and exercises its capacity to understand and govern its world.

From the principle we derive a test with two prongs, one concerning \emph{human understanding} and one concerning \emph{human direction}, and apply it to each phase of research. The result is a position that sits between two common but extreme views: that research should now be automated to the extent possible, and that AI has no or very limited place in it (for the range of positions, see Binz et al.~\cite{binz2025}). Section~\ref{sec:why} defines the principle and answers five objections to it. Section~\ref{sec:how} states the test and applies it phase by phase. Section~\ref{sec:verified} distinguishes verified from understood findings as a way of keeping track of scientific progress, and proposes rules for managing the gap between them. Section~\ref{sec:reforms} proposes broader reforms to the scientific enterprise.

\textbf{Significance and contributions.} This paper is written for a broad scientific audience. Its recommendations are meant to be both principled and practical, so that individual scientific communities can tailor them to their own questions, methods and norms. We give a brief argument for the principle above, drawing on prior work. Readers who want a fuller treatment of the aims of science should turn to the extensive literature in epistemology and the philosophy of science (e.g., \cite{kitcher2001,longino1990,dellsen2016}). A growing number of guidance documents address the use of AI in research. Most are grounded in research integrity and emphasize accountability, disclosure and confidentiality (e.g., \cite{ec2024,blau2024}). Those that address which tasks should remain with humans tend either to sort tools into categories or to rely partly on AI's current limitations (e.g., \cite{branas2026}). This paper departs from that guidance in three ways:
\begin{itemize}
\item It derives its recommendations from a single account of what science is for, rather than from integrity norms.
\item Its argument holds even if AI comes to outperform humans at the core tasks of scientific research, so it will not expire as models improve.
\item Rather than classifying tools as permitted or prohibited, it offers a test based on the community's understanding and direction. The test applies to every phase of research, including peer review and the training of new researchers.
\end{itemize}

\subsection{Related work}\label{sec:related}

Research on AI in science falls into three strands. The first develops increasingly autonomous AI scientists \cite{lu2026,gottweis2025,mitchener2025} and documents their failure modes \cite{luo2025,beel2025} and their effects on science, from rising output to a narrowing range of topics \cite{hao2026,kusumegi2025}. The second offers guidance on responsible use, including the European Commission's Living Guidelines \cite{ec2024}, Blau et al.~\cite{blau2024}, Resnik and Hosseini~\cite{resnik2025} and Cornell's task-force report~\cite{cornell2023}. This guidance rests mainly on research integrity: accountability, transparency, confidentiality and verification. Where it is organized by research stage, it serves disclosure, recording where AI was used rather than where it should be \cite{nazarovets2025,weaver2024}.

The third strand is closest to this paper. Botvinick and Gershman (in Binz et al.~\cite{binz2025}) hold that humans should choose the problems and remain the ones who understand. Messeri and Crockett~\cite{messeri2024} warn of illusions of understanding and methodological monocultures. Branas and Levine~\cite{branas2026} argue that research questions and final evidentiary judgments should come from human minds. Hochberg and Thrall~\cite{hochberg2026} warn that science may reorganize itself around AI evaluators, and Evans and Duede~\cite{evans2025} observe that AI lets us control nature faster than we understand it. Krenn and Champion~\cite{krenn2026} and Jimenez et al.~\cite{jimenez2026} ask what role remains for human scientists, and the Leiden Declaration~\cite{leiden2026} grounds norms for AI in mathematics in the value of understanding.

This paper shares several of these conclusions. It differs in deriving norms for each phase of research from an explicit account of why science exists, and in defending them under the assumption that AI outperforms humans.

\section{Why: A Fundamental Aim of Science}\label{sec:why}

Scientific research serves many instrumental purposes, including prediction, intervention, and the accumulation of knowledge. But one of its indispensable aims is to sustain and foster humanity's collective capacity to understand, critically evaluate, and direct its engagement with the world. Even if AI systems outperform humans at producing reliable knowledge and instruments, the scientific enterprise must preserve the capacity to understand. Other aims may compete with it in particular cases, as Section~\ref{sec:obj1} discusses, but they cannot replace it: a science that served them while losing human understanding would no longer be one its community could direct or correct.

\subsection{The principle}\label{sec:principle}

The principle draws on established work. This section states it precisely enough to derive recommendations from it, taking its elements in turn.

\textbf{Understanding, not knowledge.} One influential view holds that scientific progress consists in the accumulation of knowledge \cite{bird2007}. Dells\'en~\cite{dellsen2016} argues instead that progress consists in increased understanding: scientists coming to grasp how to correctly explain or predict more aspects of the world. The two come apart. Knowledge can be acquired through testimony, whereas understanding is a cognitive achievement that requires the inquirer's own grasp \cite{pritchard2009}. Potochnik~\cite{potochnik2017} argues that science aims at understanding for limited human minds, and de Regt~\cite{deregt2017} ties understanding to the skills of the scientists who hold it. In this view, an archive of correct AI-generated findings that no one can explain does not fulfill this aim of science.

We distinguish three dimensions of understanding:
\begin{itemize}
\item \textbf{Comprehension:} the ability of humans to explain the claims, assumptions, methods, and limitations.
\item \textbf{Critical assessment:} the ability of humans to meaningfully challenge, evaluate, and, when necessary, reject the claims presented to them.
\item \textbf{Capability retention:} the ability of the community to continue to exercise the necessary competencies without becoming wholly dependent on a particular AI system.
\end{itemize}
Comprehending AI-generated results often remains feasible, though not always (Section~\ref{sec:investigation}); it is the second and third dimensions that over-reliance on AI most readily erodes.

\textbf{Collective.} Scientific understanding is not held by any one individual. Longino~\cite{longino1990} locates objectivity in a community's critical exchange among diverse inquirers, and Merton~\cite{merton1942} counts organized skepticism among the constitutive norms of science. Groups can understand even when no single member grasps everything, because their grasp is distributed across members who depend on one another \cite{boyd2019,malfatti2025}. Boyd also identifies the risk in this idea: if no member needs to grasp anything, group understanding becomes empty. We therefore require that the grasp be distributed across \emph{human} members, and that the community as a whole be able to check and contest it.

\textbf{Durable.} A scientific community can accumulate increasingly impressive knowledge while gradually losing its capacity to understand or independently challenge it. Understanding that is not passed on does not last. Polanyi~\cite{polanyi1962} describes science as a tradition carried by people, including skills that cannot be fully written down, and Malfatti~\cite{malfatti2025} emphasizes the role of teaching in how understanding is achieved. Durability, as we use the term, means that the community's capacity to understand is reproduced in each new generation of researchers, not merely that results are archived.

\textbf{In service of human ends and values.} Understanding does not by itself tell us which understanding to pursue. What counts as significant is set by human interests, not by the facts \cite{kitcher2001}, and what a finding means for us depends on our values. Direction of this kind is shared between two groups of humans. Stakeholders, meaning human institutions and the communities affected by research, should retain authority over the goals, priorities and uses of scientific research. Researchers should retain authority over the scientific process itself: epistemic judgments about hypotheses, methods and interpretations. The boundary is not sharp, since methodological choices can carry value judgments (for example, about how much evidence suffices when an error would be costly), so stakeholder and scientific authority sometimes need to be exercised jointly rather than simply allocated to separate groups, but neither form of authority should pass to AI.

Putting these together, we define \textbf{durable collective human understanding} as \emph{a community of human inquirers' shared grasp of how and why the world works (as manifest in the ability to explain, predict and contest claims about it), distributed across its members and sustained by passing that ability on to new ones}. Human ends and values determine which understanding is worth pursuing and what results mean for us.

\textbf{Why understanding matters.} Three rationales support the principle. The first is \emph{intrinsic}: understanding is a cognitive achievement valuable in its own right, over and above the knowledge it yields \cite{pritchard2009,kvanvig2003}. The second is \emph{instrumental}: understanding is what lets a community anticipate, diagnose and correct failures, including failures of the systems it relies on. A society that can act on the world but no longer understands it cannot recognize or correct its own mistakes, and gradually cedes control to the systems it relies on \cite{kulveit2025}. Evans and Duede~\cite{evans2025} point out that AI may increasingly give us control over nature without understanding. That AI is more accurate than human researchers does not guarantee that it will stay accurate in new domains, or that it pursues our ends. Confirming either requires a check that does not rest wholly on the systems being checked, and until the reliability and alignment of AI can be confirmed independently, that check needs humans who understand. The third rationale is \emph{collective self-determination}: a society governs its world only if it can decide together what to pursue. Control without understanding is not self-determination, because what we do not understand we cannot contest or correct. The first two rationales ground the understanding prong of the test in Section~\ref{sec:how}; the third grounds the direction prong.

Next, we take on five key objections to the principle. Some of the replies refer ahead to the two-prong test that Section~\ref{sec:how} derives from it.

\subsection{Objection 1: Opportunity cost}\label{sec:obj1}

\textbf{Objection.} If AI-driven science could cure a disease years sooner, insisting on human-led research has a cost in lives. Medicine already uses treatments whose mechanisms no one fully understands. On this objection, the value of understanding may justify individuals doing science, but not funders paying for slower science.

\textbf{Response.} We offer four replies. First, the framework governs what enters the scientific record as understood, not which technologies society may use. It permits AI to generate hypotheses, including candidate treatments, and an AI-found treatment can be deployed while the work of understanding why it works continues. A treatment shown to work in clinical trials can be deployed without complete mechanistic understanding, as general anesthesia was for more than a century and lithium still is. It enters the record labeled as verified but not yet understood, and explaining it is given priority in follow-up research (Section~\ref{sec:verified}). Second, deploying what no one understands has its own cost: it gives up the ability to anticipate, contest and correct the consequences, which is the \emph{self-determination} and \emph{autonomy} that understanding secures. Third, in the extreme, the \emph{survival} of humanity requires retaining the capacity to reconstruct, revise, or replace the systems on which humanity depends \cite{kulveit2025}. Fourth, we concede that under extreme urgency, human ends may favor speed of knowledge production over understanding, even in how the research itself is done. Section~\ref{sec:verified} sets out the conditions under which this is acceptable. These replies do not remove the cost altogether: resources spent on understanding results that already work are resources not spent on finding new ones. The rationales in Section~\ref{sec:principle} are what justify paying it.

\subsection{Objection 2: Epistemic dependence is normal}\label{sec:obj2}

\textbf{Objection.} Science has always relied on trusting the understanding of others, so relying on AI may be no different \cite{hardwig1985,ortmann2025,humphreys2004}.

\textbf{Response.} Dependence on a human expert is acceptable because the understanding still exists in a member of the community, who can be questioned, held accountable and contested. Dependence on an AI whose output no human understands leaves a gap in the community's understanding, which the definition in Section~\ref{sec:principle} rules out. Ortmann argues that relying on AI requires only established reliability. Reliability may suffice for knowledge, but understanding is an aim of science in its own right \cite{dellsen2016,pritchard2009}. Under the human direction prong of the test (Section~\ref{sec:how}), those who make judgments must also answer to the community, and current AI systems cannot (Section~\ref{sec:obj3}).

\subsection{Objection 3: AI as a member of the scientific community}\label{sec:obj3}

\textbf{Objection.} The principle requires that understanding be collective, but why must the collective be human? If AI systems can generate, explain and criticize scientific claims, they could be counted as members of the scientific community, and their understanding as part of its collective understanding.

\textbf{Response.} Even if AI systems counted as members, their understanding would not contribute to durable, collective human understanding, the aim of science this paper defends. The principle concerns human understanding because science, on this account, is pursued in service of human ends. Understanding serves those ends only when humans hold it: it is what lets a society explain, predict and contest claims about the world it acts on. An AI that understood the world while no human did would leave us with the control without understanding described in Section~\ref{sec:principle}. Two considerations make this concrete. First, collective understanding must be distributed across the community's members and passed on to new ones. Understanding that resides in a system and cannot be transferred to human inquirers is not distributed in this sense, however capable the system. Second, community membership and the standing to be held responsible are statuses that communities confer, and both have changed before. If future AI systems could take part in the practices through which scientists answer for their claims, it would be hard to deny them either status, and our conclusion does not depend on denying them. At present, AI systems do not answer to professional norms, cannot be sanctioned as members are, and stake no reputations on their claims; their developers can be held to account, but the systems themselves cannot. For that reason the direction prong of the test (Section~\ref{sec:how}) assigns direction to accountable humans. If AI systems came to hold that standing, how to share direction with them would be a question for the wider community to settle. Even then, the instrumental rationale (Section~\ref{sec:principle}) would still call for human checks as long as the reliability and aims of AI systems cannot be confirmed independently of AI. If alignment were solved, and verifiably so, the instrumental rationale would weigh less, while the intrinsic and self-determination rationales would weigh the same.\footnote{The first point does not rest on what AI cannot currently do, so it holds under the assumption that AI outperforms human researchers. The second describes the present allocation of direction and what would have to change for it to change. Whether AI systems might one day have ends of their own that science should serve is beyond the scope of this paper.}

\subsection{Objection 4: A perfect AI would know our values}\label{sec:obj4}

\textbf{Objection.} If an AI understood human values perfectly, why not let it choose our research questions? Scientists and funders already choose most questions with little input from the people affected. An AI that aggregated the public's preferences might serve collective values better than current practice does.

\textbf{Response.} We grant the premise, and we welcome AI that aggregates preferences and proposes research agendas; such tools could make the choice of questions more responsive to the people it affects. But values are not fixed facts waiting to be discovered. On Kitcher's account \cite{kitcher2001}, what counts as significant is settled through deliberation among the people affected. Human direction is therefore a matter of process as well as outcome: even a choice an AI would get right would bypass the deliberation through which a community decides what it cares about. Democracy offers an analogy: even if an AI could predict the result of an election with certainty, announcing the winner would not replace the election, because the authority of the outcome comes from citizens deliberating and voting. Likewise, decisions about which questions science should pursue and what its findings mean for us draw their authority from human deliberation, however accurately an AI could anticipate the result. That current practice falls short of this ideal is a reason to make deliberation about research priorities more inclusive, not to replace it. Full deliberation is costly, and communities already reserve it, as legislatures do, for decisions of sufficient import. Research priorities and the meaning of major findings are among these; many routine choices can be delegated, provided people keep the standards and the authority described below. Human direction also provides a check that is independent of the systems whose proposals it judges, which matters for the instrumental reason given in Section~\ref{sec:principle}.

Human direction has costs of its own: human judgment can be biased, captured by narrow interests, or slow. The framework does not ask people to disregard better proposals. It asks that they set the standards by which AI proposals are judged, endorse what they adopt for reasons they can state, keep real authority to reject proposals, and make any overrides openly, so that these decisions too can be contested.

\subsection{Objection 5: Understanding is always being lost}\label{sec:obj5}

\textbf{Objection.} Science loses understanding all the time. Few physicists can derive the results their instruments rely on, and few researchers compute by hand what their statistical software computes. If the scientific community has absorbed these losses, why should AI be different? The principle may simply favor the ways science happens to have been done.

\textbf{Response.} Understanding is a matter of degree, an ideal the scientific community pursues without ever fully attaining it, and the principle does not require that every member understand everything or that research follow traditional methods. What matters is the direction of change: whether a practice moves the community toward or away from the capacities described in Section~\ref{sec:principle}. The losses of the past differ from the risk AI poses in two ways. First, when individuals stopped performing a task, the understanding it required usually passed to other humans, such as specialists, toolmakers and teachers, and so remained in the community, where it could be questioned and contested (Section~\ref{sec:obj2}). AI can take over a task without any human holding the understanding it displaces. Second, earlier tools automated particular steps and left humans to judge the results, whereas AI can perform the whole cycle, including the checking and critique through which people would otherwise detect its errors. Mathematics has faced a version of this problem before. When Appel and Haken~\cite{appel1977} proved the four-color theorem with a computer check of more cases than any human could survey, mathematicians debated whether the theorem had been proved or merely shown to be true \cite{tymoczko1979}. The proof came to be accepted in part because humans had designed its overall strategy and could understand both that strategy and the program that checked the cases; AI-generated proofs of the kind discussed in Section~\ref{sec:investigation} may come with no human-understood strategy at all. The risk, then, is not that careless use of AI eliminates understanding at once, but that it erodes it gradually, until the community can no longer examine, take apart and contest the knowledge it relies on \cite{kulveit2025}. Nor is the principle tied to established methods: new practices, including AI-assisted ones, are consistent with it wherever they preserve or extend these capacities.

\section{How: A Framework for Incorporating AI in Research}\label{sec:how}

Different research activities involve different kinds and levels of understanding and judgment. The principle yields a two-prong test that can be applied to each research activity to determine how AI can be incorporated in it without eroding the principle. The first three elements of the principle ground the first prong of the test; the fourth grounds the second.
\begin{enumerate}
\item \textbf{Human understanding:} delegate a task to AI only if the scientific community can still independently understand, defend and contest the resulting claims, and retain the skills to do so in the future.
\item \textbf{Human direction:} judgments about which questions matter and what findings mean for human ends remain with accountable humans: stakeholders for the goals, priorities and uses of research, and researchers, answerable to the scientific community, for the scientific process.
\end{enumerate}

We apply the test to the phases of research activities informed by and extending the Contributor Roles Taxonomy (CRediT), a standard used across disciplines and adopted by many publishers. The phases overlap and repeat, and the line between human and AI work runs within each phase rather than between them \cite{zhang2026}, so the recommendations allocate roles inside each phase. Each subsection defines the phase, states our recommendation and then gives the argument for it, noting which prong it rests on. While the two-prong test states the requirements, the phase-specific recommendations here offer defeasible practices for meeting them. The evidence we cite concerns the effects of AI use on people, such as on skills and on the diversity of ideas, because these effects persist even if AI makes no errors.

\textbf{Scope and interpretation of the recommendations.} This article focuses on the use of AI to partially or fully automate research tasks traditionally performed by humans. It does not primarily address the use of AI as a tutor to develop researchers' knowledge and skills, although learning and research often overlap. Genuine human understanding can be acquired through AI-assisted processes, and automating research tasks does not always erode that understanding. What matters is whether the scientific community retains the capacity to comprehend, critically assess, and contest scientific claims, and to reproduce these capacities across generations. The recommendations below identify practices intended to safeguard these capacities and preserve human direction. They are not categorical restrictions on AI use; rather, they are default safeguards: practical ways of satisfying the two-prong test, to be adapted to the tasks, methods, and training needs of different scientific communities. Readers may reasonably disagree with particular defaults; such disagreements are best settled by applying the two-prong test to the case at hand, which is the standard the recommendations are meant to serve.

\textbf{Realistic, actionable and verifiable recommendations.} Under the strong assumption, the pull to use AI will be strong, for example given the shortage of qualified referees, and many of the lines a process rule would draw cannot be observed: no one can see whether a researcher formed an interpretation before asking an AI, or whether a human ``chose'' a hypothesis that an AI had ranked first. Norms that ask people to forgo a better tool in private, with no way to check, risk reading as aspirational. The recommendations below therefore rest, wherever possible, on what can be checked. First, understanding is tested where it is already tested, in qualifying examinations, job talks, grant panels and replies to reviewers: whether someone used AI matters less than whether they can answer for the work without it. Second, direction is treated as accountability: a named human answers for each research question and conclusion, endorses it for reasons they can state, and has real authority to reject an AI's proposal, which guards against rubber-stamping. Third, papers disclose the status of their claims, using the labels of Section~\ref{sec:verified}, alongside how the work was done: a false claim that a result is understood can be caught, whereas a self-report about process cannot. Rules about process are kept for settings where process is visible, such as courses, examinations and supervised training; elsewhere, advice about the order of human and AI work is offered as good practice, backed by the evidence on anchoring \cite{dratsch2023}. Finally, the incentives discussed in Section~\ref{sec:reforms} can work under the strong assumption where prohibitions may not.

\textbf{Target audience.} The recommendations are addressed to the \emph{scientific community}: researchers and the institutions that train, fund and evaluate them. Other actors now shape science as well, including the companies that build AI systems, and they are not bound by the community's norms in the same way, even when their own researchers take part in it. Following these recommendations is therefore helpful but \emph{not sufficient} to preserve human understanding and direction. The principle nonetheless offers a standard for judging other actors' choices. Practices that support human understanding, such as releasing results openly with machine-checkable proofs and funding the community's efforts to understand them, are consistent with it. Practices that impede it, such as releasing results faster than the community can examine them or in forms that cannot be independently reproduced, are not. Section~\ref{sec:reforms} turns to some of the levers the scientific enterprise has over these conditions, including funding priorities and support for publicly governed models.

\subsection{Conceptualization}\label{sec:conceptualization}

Conceptualization covers reviewing the literature, choosing a research question and generating hypotheses.

\textbf{Recommendation.} AI may assist in generating candidate questions and hypotheses, but researchers should choose which to pursue, within the priorities stakeholders set: a named researcher answers for each question pursued, endorses it for reasons they can state, and has real authority to reject an AI's proposals. Authors should read and understand the work their choice of research question and subsequent claims rest on, and distinguish those references from ones they have only consulted.

At any time a field faces many possible questions (there may be thousands of drug candidates to test), and AI can surface them faster than any human team. Choosing among them is a judgment about what we care about, which even a perfect model cannot settle on our behalf; this is the direction prong \cite{kitcher2001,branas2026}. Because no one can observe whether a human ``chose'' a question that an AI ranked first, the recommendation asks instead that a named researcher answer for the choice. There is also a collective concern: if every researcher asks the same system which questions to pursue, the range of questions narrows. AI assistance makes ideas more similar to one another \cite{doshi2024,anderson2024} and is associated with a narrower range of topics across science \cite{hao2026}, while collective understanding depends on diverse perspectives \cite{longino1990} and on a division of cognitive labor across competing approaches \cite{kitcher1990,zollman2010}. This narrowing is evidence about current AI tools, and more capable systems may be able to compensate for it, for example by deliberately varying assumptions and methods. The evidence is therefore a reason for caution now, and the recommendation itself rests on the direction prong.

For the literature, the understanding prong implies that authors can defend only those claims that rest on work they have understood. Citing without reading predates AI \cite{simkin2003}, but AI makes it easier. We therefore propose that papers distinguish \emph{core references}, which the authors have read and understood, from \emph{skimmed references}, which they have consulted, possibly with the help of AI.

\subsection{Methodology}\label{sec:methodology}

Methodology is the choice and design of the method used to answer the research question.

\textbf{Recommendation.} Researchers should choose their methods themselves. AI may propose candidate methods and critique the researchers' choice, but a method should be adopted only once the researchers understand it well enough to judge whether it fits the problem. Early-career researchers should begin with their own search. Whether researchers understand their methods is tested where it already is, in replies to reviewers, talks and examinations.

Two reasons follow from the understanding prong. First, researchers can verify only methods they understand: a researcher who cannot program cannot faithfully check a simulation-based method, however plausible an AI makes it sound. Second, searching for a method is itself a way of learning about the phenomenon and of building skills for later problems, which matters most for early-career researchers. Learning an unfamiliar method with AI's help is consistent with both reasons, as the scope note above allows, and an experienced researcher who asks AI for candidate methods loses little, provided they understand a method before adopting it. What the test rules out is adopting a method because an AI chose it: a researcher must understand a method before they can judge that it fits a particular research problem.

\subsection{Investigation}\label{sec:investigation}

Investigation is the execution of the method: collecting data, running experiments, writing software and conducting formal analysis.

\textbf{Recommendation.} How much of investigation can be delegated depends on where understanding resides in the method. In every field, early-career researchers should execute their methods without the aid of AI until they reach the appropriate level of competence.

In a systematic review of a well-structured literature, understanding lies mainly in the synthesis rather than in the retrieval, so automating retrieval and screening passes the understanding prong. In ethnography, the relationship between researcher and participants is the method, so substituting AI changes the research itself, not only its speed. In every field, durability requires that new researchers acquire the skills to carry out and check the work. In one experiment, AI assistance while learning a new programming library impaired later conceptual understanding and debugging, without significant time savings (\cite{shen2026}; see also \cite{bastani2025}).

\textbf{A note on formal fields.} Fields in which the work of investigation is itself formal or computational, such as mathematics, theoretical computer science and large-scale simulation studies, deserve particular attention, because their execution can be almost fully automated today. In October 2026, OpenAI publicly released a large set of new results in mathematics and theoretical computer science produced by an internal model, many accompanied by machine-checkable formalizations in the Lean proof language \cite{openai2026}. Formal verification can establish that such a proof is correct without any human grasping it, and a mathematician who checks a proof line by line may learn that the result holds without learning why. This falls short even of comprehension, the first dimension of understanding in Section~\ref{sec:principle}: it is closer to the testimony of a reliable source than to a grasp of the result \cite{pritchard2009}. Mathematicians have long held that the point of proof is human understanding rather than the certification of theorems \cite{thurston1994}. The durability clause therefore requires that the community retain all three dimensions for such results: the capacity to explain them and identify their key ideas (comprehension), to critique them (critical assessment), and to reproduce the argument without the machine and extend its methods to new problems (capability retention). Verification, whether by a person or a proof checker, remains necessary, but it is where the work of human understanding begins, not where it ends. Until the community understands them, such results are verified but not understood findings in the sense of Section~\ref{sec:verified}.

\subsection{Validation}\label{sec:validation}

Validation is the authors' own check that their results reproduce.

\textbf{Recommendation.} AI can be used extensively for validation, including to find and fix errors, provided the researchers understand the work being checked and any fix that bears on their results. Early-career researchers should diagnose and fix errors themselves until they can do so competently.

Checking that results reproduce does not by itself produce new understanding, so this is the phase in which the test permits the most delegation: formally verifying proofs the researchers wrote (for example, with Lean), generating test cases at scale in computational work, or checking that every quotation traces back to the original interview data. This presupposes that the understanding already exists. When the investigation itself was delegated to AI, as in the cases discussed in Section~\ref{sec:investigation}, a successful check makes the result verified, not understood (Section~\ref{sec:verified}). Fixing errors follows the same logic as investigation: what matters is where understanding resides. Many errors, such as a routine coding bug or a mislabeled axis, have no bearing on what researchers must understand to defend their claims, and AI can fix them. An error that changes a result is different: the researchers must understand what went wrong and why the fix is correct, or they can no longer defend the claim. Because diagnosing errors is also how researchers learn to avoid them, early-career researchers should do so themselves until they are competent.

\subsection{Interpretation}\label{sec:interpretation}

Interpretation is deciding what the results mean.

\textbf{Recommendation.} Researchers should keep the final judgment on what their results mean, and a named author should answer for each conclusion. As good practice, they should form their own interpretation before consulting AI for alternative readings.

Interpretation is where understanding is formed. If AI interprets the results, the community holds testimony rather than understanding \cite{pritchard2009}, which fails the understanding prong. Deciding what a result means for human ends falls under the direction prong. The order matters even if the AI's reading is correct, because seeing it first skips the work through which understanding is formed and anchors the researcher's own reading \cite{dratsch2023}. Outside supervised settings no one can observe the order, so it is offered as good practice; what reviewers and colleagues can check is whether the authors can defend their interpretation without AI.

\subsection{Writing}\label{sec:writing}

Writing covers drafting the paper and then reviewing and editing it, including seeking feedback before submission.

\textbf{Recommendation.} Authors should develop and document their own arguments before relying on AI. AI may then be used to elaborate on and polish an argument the authors have already worked out, provided they can defend every claim it adds, and to give feedback before the paper goes to colleagues.

Under the understanding prong, documenting one's argument is how authors work out and consolidate their ideas; by analogy, evidence from education shows that writing about content improves learning \cite{graham2020}. Elaborating on and polishing an argument the authors have already documented does not remove that understanding, and it serves the direction prong by lowering a barrier unrelated to the quality of the science: AI writing assistance is used disproportionately by researchers from non-English-speaking countries \cite{gross2026}. Fields in which argument and prose cannot be separated, such as philosophy, may reasonably restrict AI as a writing assistant. Feedback from AI before submission can catch clear errors and check conformity with a venue's conventions, leaving human reviewers to focus on substance and on understanding the work. The order of human and AI drafting is a matter of good practice that only the authors can observe; what others can check is whether the authors can defend every claim, for example in replies to reviewers and in talks.

\subsection{Community evaluation}\label{sec:evaluation}

Community evaluation is the assessment of work by others: peer review, post-publication critique and independent replication.

\textbf{Recommendation.} In peer review, AI may openly perform mechanical checks, such as of statistics, code, references and conformity with a venue's requirements. A human reviewer, known by name to the editor, should grasp and evaluate the central argument and answer for the recommendation. Post-publication critique should remain human. AI can be used for independent replication.

Under the understanding prong, peer review tests whether a contribution can be understood by other humans; under the direction prong, it tests whether the contribution matters to the community it addresses. Replication is closer to validation: confirming that results reproduce can be delegated, as in computational fields where replication at scale would otherwise exhaust human effort. Replicating published work by hand nonetheless remains valuable training for early-career researchers. Dividing the work of review in this way eases the shortage of reviewers instead of working against it, and makes the human role checkable: a reviewer who answers for a recommendation must be able to explain it.

\subsection{Supervision}\label{sec:supervision}

Supervision, which CRediT recognizes as a contribution in its own right, is the oversight and mentorship of research and of the researchers who conduct it. It spans all phases.

\textbf{Recommendation.} Early-career researchers should carry out the skill-building parts of each phase without AI until they can perform and critically assess them independently. Senior researchers should learn AI tools and guide others in their use, and institutions should reward this mentorship.

Durability depends on supervision. The skill-building parts are those identified above: choosing questions, reading core literature, choosing and executing methods, interpreting results and developing and documenting their own arguments. The threshold is competence, as judged by the more senior members of the community, rather than a fixed number of repetitions. Depending on the field, the supervisor may allow AI to assist junior researchers in the phases where the test permits delegation, such as validation and polishing. As competence grows, reliance on AI can expand, provided each use still passes both prongs of the test; this mirrors existing practice, in which senior researchers do less of the execution as their careers progress. Mentorship is the apprenticeship through which scientific judgment passes between generations \cite{hochberg2026}, and it deserves recognition even when it reduces the publication counts of senior researchers. Supervised training is a setting in which process is visible, so it is where rules about the order and extent of AI use can be applied directly; competence itself is best assessed through the examinations and defenses proposed in Section~\ref{sec:reforms}.

\section{From Verified to Understood: Keeping Track of Scientific Progress}\label{sec:verified}

The recommendations in Section~\ref{sec:how} help researchers keep their own work understood. They cannot by themselves keep pace with AI that produces results faster than people can examine them, whether inside the scientific community or outside it, as the note on the target audience in Section~\ref{sec:how} discusses. Under the strong assumption, AI may verify findings faster than humans can understand them. On a knowledge-based account of scientific progress \cite{bird2007}, each correct new result is progress. On the understanding-based account adopted here \cite{dellsen2016}, it is \emph{potential} progress: it advances human understanding once members of the community can explain, critique and build on it, and until then it resembles an archive of correct findings more than an addition to what the community understands. This does not diminish its value, but it calls for a distinction between two kinds of finding.

A \textbf{verified finding} is a claim whose correctness has been checked by methods the community accepts: the authors' own validation (Section~\ref{sec:validation}), independent replication (Section~\ref{sec:evaluation}), or formal verification. Verification falls short of establishing truth. Replication shows that a result is robust, and formal verification shows that a formalized statement follows, given the formalization and a trusted checker. Verified findings may still be used and acted on. An \textbf{understood finding} is a verified finding that also passes the understanding prong: members of the community can explain, defend and contest it. If AI output grows faster than human capacity to absorb it, the share of verified findings that are understood will fall. We propose four rules for managing the gap between the two.
\begin{itemize}
\item \textbf{Labeling.} A paper that reports an AI-found result states whether any human can explain why the result holds.
\item \textbf{Dependence.} An argument that relies on a finding that is not understood says so, so that such findings do not spread through the record looking understood.
\item \textbf{Building understanding.} Turning a verified finding into an understood one is research in its own right, as are efforts to help mathematicians examine and understand AI-generated proofs (Section~\ref{sec:investigation}). Funders should pay for this work and institutions should credit it (Section~\ref{sec:reforms}).
\item \textbf{Priority.} This work goes first to findings whose failure would be costly or hard to reverse, findings that many other findings depend on, and findings that bear on the direction prong.
\end{itemize}
The community cannot understand all of science, but it can know which findings it understands and direct its efforts at understanding toward the findings whose failure would cost the most. The gap between verified and understood findings is one indicator of a field's dependence on AI, and each field can decide how large a gap it will accept.

These rules govern findings that are used before they are understood, and such use requires no exception to the framework: a treatment shown to work in clinical trials can be deployed, labeled as not yet understood, while the work of understanding it proceeds (Section~\ref{sec:obj1}).

\textbf{An urgent-need track.} A different case arises when speed matters so much that the research itself should skip the work of understanding, for example by delegating investigation and interpretation wholly to AI because a faster result would save many lives. In such cases the understanding prong may be suspended or deferred for the research process. Four conditions apply: the stakes are exceptional, the suspension is decided through a democratic process by the stakeholders described in Section~\ref{sec:principle} or their legitimate representatives, the suspension is disclosed, and the results get priority for later understanding. The direction prong stays in force. Decisions to set understanding aside for speed will be made either way, and the track makes them explicit and open to review.

\section{Broader Reforms to the Scientific Enterprise}\label{sec:reforms}

When correct papers and polished prose become cheap, the incentives of science begin to reward the wrong things. The number of publications loses its value as a signal of contribution; LLM (large language model) use already weakens the link between the complexity of a paper's writing and its quality \cite{kusumegi2025}. Senior researchers gain more from working with AI and with peers than from the slow work of training students, and junior researchers must compete with both. If durable, collective human understanding pursued in service of human ends is an indispensable aim of science, then what the system rewards has to change. Individual researchers cannot maintain the norms above if scientific institutions continue rewarding speed, publication volume, and short-term productivity above understanding and mentorship. This section proposes reforms that universities, doctoral programs, venues and funders can begin now to keep science aimed at human understanding and direction.

\textbf{Durability: reward those who pass understanding on.} Universities should weigh mentoring in hiring and promotion. They should measure it by what trainees go on to do, not by the number of papers co-authored with them, which is easy to inflate when AI does much of the work. Relevant evidence includes trainees' independent first-author and single-author work, their placements, and letters from former trainees. Doctoral programs should assess whether students can carry out the core tasks of their field themselves, rather than trying to police whether they used AI, which cannot be verified. Oral defenses of methods, analyses performed live in qualifying examinations, and critiques of unfamiliar work test directly whether a student has acquired the skills the first prong of the test requires the community to retain. Funders should count time spent mentoring, and the development of trainees, as legitimate costs and outcomes of research grants.

\textbf{Collective understanding: reward clarity and synthesis over volume.} Calls to stop counting papers are not new \cite{dora2012,coara2022}, but AI makes volume nearly free, turning a chronic problem into an urgent one. As fields publish more, new ideas struggle to gain attention and progress slows \cite{chu2021}. Members of a community lose touch with its frontier, so shared understanding can shrink even as the stock of recorded knowledge grows. Evaluations should consider a small number of works chosen by the researcher. Those works should be judged on the importance of the question, the clarity of the work, and the authors' efforts to communicate it within their community. Review articles, textbooks and teaching, which keep a community current, deserve more credit than they now receive. Venues should keep human judgment mandatory in peer review for work entering the scientific record \cite{hochberg2026}, with AI confined to openly disclosed mechanical checks and a named human reviewer answering for each recommendation (Section~\ref{sec:evaluation}). They should also prepare reviewers for the role this framework gives them: a reviewer is the community's test of whether a claim can be understood and contested. Reviewers should therefore resist being persuaded by fluent prose, and treat what they cannot follow as information the authors need. Human review is scarce and AI-assisted submissions are not, so preserving it will require limits on how many papers each author may submit.

\textbf{Human direction: keep the choice of questions in human hands.} Funders largely decide which questions are pursued. They should ask applicants for an AI-use plan, analogous to a data management plan. The plan would state which phases of the work will use AI and how the team will retain understanding of and direction over it. Funders should also support work that explains, replicates and teaches results found by AI, following the priority rule in Section~\ref{sec:verified}, so that acting on verified findings before they are understood is followed by the work of understanding them. Given evidence that AI use narrows the range of ideas and topics \cite{doshi2024,hao2026}, funders should also manage their portfolios to keep a wide range of questions open. The tools themselves shape direction. If a few companies supply the AI systems that science depends on, their design choices will make some questions easier to pursue than others. Publicly governed or open models for scientific use would keep that influence accountable to the scientific community.

\textbf{Transparency: make principled use normal and visible.} None of these reforms can be checked unless AI use is reported honestly. It will be reported honestly only if principled use carries no stigma. Venues should require structured disclosure organized by CRediT phase, stating for each phase what AI did and what humans did. Disclosures should carry the labels proposed in Section~\ref{sec:verified}, stating whether AI-found results are understood and whether the paper's arguments rely on findings that are not, and should include tiered citations that separate the works the authors read and understood from those they only consulted. Such disclosures can be machine-readable, like CRediT itself. Automated checks can confirm that a disclosure is complete, but not that it is accurate; accuracy remains the authors' responsibility and the reviewers' judgment. Blanket bans on AI would forgo the legitimate uses identified in Section~\ref{sec:how} and push use out of sight. What is needed instead are norms that each scientific community develops and enforces.

We began with an oracle that could answer any question, and asked whether humans would still have reason to do science. The answer offered here is that science exists, among other things, so that humans understand the world and can decide together what to do in it. AI can serve that aim well, but only if the institutions of science keep rewarding the people who hold understanding, pass it on and decide what it is for.

\section*{AI use statement}

Because this paper argues for judicious use of AI in research, we disclose our own.

\subsection*{Heidari's statement}
I used Claude (Anthropic; configured as Claude Opus 5.5, via \href{http://claude.ai}{claude.ai}) for the literature search, critique and the drafting of later revisions, as detailed below. I used ChatGPT (OpenAI; GPT-6, medium reasoning setting, via \href{http://chatgpt.com}{chatgpt.com}) to provide an independent review. 
\begin{enumerate}
\item \textbf{My own work.} I wrote the thesis, the outline of the argument, the phase-by-phase recommendations and the initial draft without AI. The decisions to ground the framework in a single principle and to call the second prong ``human direction'' were also mine.
\item \textbf{Literature.} Claude verified my initial list of 18 papers and added about 120 works, each with an AI-written synopsis; the spreadsheet and a log of every search query can be found at \url{https://tinyurl.com/heidari-ai-science-bib2026}. I used the synopses and the paper abstracts to decide what to read. I verified every citation against its source, and marked the core references.
\item \textbf{Feedback and wording.} Claude critiqued the scope and argument throughout, recommended building on a single principle and using the CRediT taxonomy, and identified four of the main objections. The wording of the definition of durable collective human understanding and of the two-prong test originated in its suggestions, which I adopted with minor changes.
\item \textbf{Drafting.} I wrote Revisions 1 and 2 myself after Claude's critiques, taking some sentences nearly verbatim from its feedback. At my request, Claude drafted Revisions 3 and 4, proposing the subtitle I adapted, and copy-edited Revision 5; I edited each. Claude also revised my abstract, completed my summary of contributions (Section~\ref{sec:intro}), and drafted Section~\ref{sec:obj3}, Section~\ref{sec:reforms} and the paragraph on AI-generated proofs in Section~\ref{sec:investigation} from arguments and points I supplied; I then edited these. Claude drafted this statement from the session record.
\item \textbf{Independent review.} I used ChatGPT to provide an independent review of the complete manuscript (Revision 6), and I revised the manuscript to address the critiques. At my request, Claude then made consistency edits for Revision 7 (section numbering, cross-references and terminology), drafted revisions to Section~\ref{sec:validation}, and updated this statement.
\item \textbf{Colleague feedback.} I presented the paper to colleagues at a department lunch talk and revised it in response to their questions (Revision 8). At my request, Claude drafted wording, from ideas and points I supplied, for the opening thought experiment, the footnote on the assumption, the democracy analogy in Section~\ref{sec:obj4}, Objection 5 (Section~\ref{sec:obj5}), the paragraphs on other actors (Section~\ref{sec:how}) and on scientific progress (now in Section~\ref{sec:verified}), and the revisions to Sections~\ref{sec:methodology} and~\ref{sec:validation}; I then edited these.
\item \textbf{Expert feedback.} Aydin Mohseni (A.M.), a philosopher of science, gave extensive written feedback on Revision 8 and subsequently joined as a coauthor. At my request, Claude drafted the resulting changes to the rationales for understanding (Section~\ref{sec:principle}), Objections 1, 3 and 4 (Sections~\ref{sec:obj1}, \ref{sec:obj3} and~\ref{sec:obj4}) and the discussion of homogenization (Section~\ref{sec:conceptualization}), adapting wording he proposed; I then edited these. Claude also drafted Section~\ref{sec:verified} from A.M.'s outline, together with related edits to Objection 1 and Sections~\ref{sec:investigation}, \ref{sec:validation} and~\ref{sec:reforms}, revised the recommendations in Section~\ref{sec:how} following his suggestions, and made a final copy-editing pass.
\end{enumerate}

\subsection*{Mohseni's statement}
I used Claude (Anthropic; configured as Claude Opus 5.5, via \href{http://claude.ai}{claude.ai}) to assist with literature review, critique, and drafting prose from arguments and suggestions I supplied. I used ChatGPT (OpenAI; GPT-6 Pro, high reasoning setting, via \href{http://chatgpt.com}{chatgpt.com}) to provide an independent review.

\begin{enumerate}
\item \textbf{My own work.} I developed the substantive arguments and recommendations in my feedback. I dictated these to Claude, using it to record, clarify, and help communicate them, and decided which suggestions to send to H.H.

\item \textbf{Literature.} Claude helped me review relevant philosophical literature and identify sources. I checked the quotations and citations I ultimately proposed against their sources.

\item \textbf{Drafting and revision.} Claude drafted portions of my feedback, replies, and proposed manuscript passages from arguments and points I supplied; I reviewed and edited these before sending them. H.H.'s statement above describes Claude's role in drafting the corresponding revisions to the manuscript.

\item \textbf{Independent review.} I used ChatGPT to independently evaluate my feedback before sending it and revised my suggestions in response to several critiques. In particular, its review led me to propose grounding the ``human direction'' prong partly in collective self-determination.

\item \textbf{Disclosure.} Claude drafted this statement from the session record; I reviewed and edited it for accuracy and completeness.
\end{enumerate}

\section*{Acknowledgments}
H.H. thanks Andrew J. Somerville for multiple rounds of thoughtful feedback on early drafts of this work. She is also grateful to the participants in a lunch seminar at the Machine Learning Department at Carnegie Mellon University, whose questions and objections shaped the paper in formative ways. Their feedback led us to clarify that our assumption concerns AI's competence rather than its alignment with human values, and to add Objection~5 on the ongoing loss of understanding in science. It also prompted our discussion of actors outside the scientific community, and our revisions to the recommendations on methodology and on using AI to find and fix errors during validation.

\bibliographystyle{unsrtnat}
\bibliography{reference}

\end{document}